\documentclass[conference]{IEEEtran}
\IEEEoverridecommandlockouts
\usepackage{amssymb}
\usepackage{url}
\usepackage{times}
\usepackage{booktabs}
\usepackage{tabularx}
\usepackage{lineno}
\usepackage{amsmath}
\usepackage{amsfonts}
\usepackage{setspace}
\usepackage[noabbrev]{cleveref}
\usepackage{graphicx}
\usepackage{subcaption}
\usepackage{float}
\usepackage{caption} 
\usepackage{makecell} 
\usepackage{siunitx} 
\usepackage{array} 
\usepackage[flushleft]{threeparttable} 
\usepackage{lipsum}

\def\BibTeX{{\rm B\kern-.05em{\sc i\kern-.025em b}\kern-.08em
    T\kern-.1667em\lower.7ex\hbox{E}\kern-.125emX}}
\begin{document}

\title{Design of a six wheel suspension and a three-axis linear actuation mechanism for a laser weeding robot\\
}

\author{Muhammad Usama, Muhammad Ibrahim Khan, Ahmad Hasan, Muhammad Shaaf Nadeem,\\Khawaja Fahad Iqbal, Jawad Aslam, Mian Ashfaq Ali, Asad Nisar Awan \\ Robotics and Intelligent Systems Engineering (RISE) Lab, Department of Robotics and Artificial Intelligence, \\ School of Mechanical and Manufacturing Engineering (SMME), \\ National University of Sciences and Technology (NUST), Sector H-12, Islamabad
44000, Pakistan
}

\maketitle

\begin{abstract}
 Mobile robots are increasingly utilized in agriculture to automate labor-intensive tasks such as weeding, sowing, harvesting and soil analysis. Recently, agricultural robots have been developed to detect and remove weeds using mechanical tools or precise herbicide sprays. Mechanical weeding is inefficient over large fields, and herbicides harm the soil ecosystem. Laser weeding with mobile robots has emerged as a sustainable alternative in precision farming. In this paper, we present an autonomous weeding robot that uses controlled exposure to a low energy laser beam for weed removal. The proposed robot is six-wheeled with a novel double four-bar suspension for higher stability. The laser is guided towards the detected weeds by a three-dimensional linear actuation mechanism. Field tests have demonstrated the robot's capability to navigate agricultural terrains effectively by overcoming obstacles up to 15 cm in height. At an optimal speed of 42.5 cm/s, the robot achieves a weed detection rate of 86.2\% and operating time of 87 seconds per meter. The laser actuation mechanism maintains a minimal mean positional error of 1.54 mm, combined with a high hit rate of 97\%, ensuring effective and accurate weed removal. This combination of speed, accuracy, and efficiency highlights the robot's potential for significantly enhancing precision farming practices.
\end{abstract}


\section{Introduction}
The world is witnessing an increasing application of autonomous devices in agriculture \cite{Hajjaj2016}. Field robots are being extensively used in agriculture due to their adeptness in navigating different terrains and performing labor-intensive tasks timely and effectively \cite{6603408}. As a result of automated farming, crop yield and quality are also increased manifold \cite{ Lowenberg-DeBoer2020}. With the trend of precision farming increasing across the globe, agricultural robots are being used to monitor crop and soil parameters, optimize resource utilization and reduce the environmental impact of farming \cite{BECHAR201694}. 

Weeding is one of the most vital aspects of farming. It is the process of removing unwanted plants and herbs from cultivated crops. Weeds not only compete with crops for nutrients and resources, but also adversely affect quality and yield \cite{SLAUGHTER200863}. The average loss of crops yields due to weeds is estimated to be 28\%\cite{Vila2021}. Traditionally, weeds are removed either by manual method such as pulling and hoeing \cite{ErgoAmjad} or by use of herbicides \cite{Kudsk_Streibig_2003}. However, both methods have their limitations in terms of efficiency and environmental impact.

Recent advancements in automated weed removal have seen the development of mobile robot-based methods \cite{SLAUGHTER200863} and drone applications \cite{Esposito2021}. To minimize herbicide usage, Raja et al. developed a high-speed, centimetre precision spray actuator system, paired with a pixel-based weeds and crops classification algorithm to spray weeds detected between lettuce plants \cite{RAJA202331}. Moreover, Hu et al. introduced a software and hardware system that employs a mobile robot for early-stage weed control, utilizing micro-volume herbicide spray. This system strategically assigns nozzles according to the sizes of the weeds \cite{9830878}. 

Though precision spraying by mobile robots results in lower herbicide consumption, it still contributes to soil degradation, environmental contamination and risk of chemical drift into water bodies \cite{Kudsk_Streibig_2003}. Herbicide usage also faces challenges due to the growing number of herbicide-resistant weeds, with 269 such species discovered as of 2022 \cite{WSSA2023}.

Non-chemical methods of weeding have also been developed. Chang et al. presented a mechanical weeding method using a modular, pyramid-shaped tool on a mobile platform, which shovels out the weeds without disturbing the soil  \cite{agriculture11111049}. In order to reduce crop damage using mechanical tools, Quan et al. developed an intelligent intra-row mechanical weeding system that utilizes deep learning for crop and weed detection and establishes protected and targeted weeding zones for tool operation in proximity to the crops \cite{QUAN202213}.  

However, mechanical weeding not only requires crops to be well distributed to avoid crop damage but also inadvertently harms beneficial organisms such as beetles and earthworms on the soil surface \cite{Earthworm2017}. It also aggravates the risk of soil erosion and leaching of plant nutrients along with triggering dormant weed seeds to germinate \cite{MechWeedCloutier}.

Given the environmental and efficiency challenges posed by chemical and mechanical weeding, there is a need to look towards novel weeding methods. In this context, laser weeding has emerged as a promising solution for sustainable agriculture. Heisel et al. demonstrated that the use of a laser beam to cut weeds stems can be an effective weed control method \cite{WeedCO2Heisel}. Integration of this technique with use of mobile robots for removing weeds through laser exposure has been gaining traction recently \cite{LWFOF}.  Marx et al. designed a weed damage model for laser-based weed control to determine minimal effective laser doses and impact of laser spot positioning for different types of weeds  \cite{MARX2012148}.

Recent advances in laser weeding involve the use of mobile robots, equipped with laser actuators, that are capable of traversing fields to accurately detect and eliminate weeds. Xuelei et al. proposed a kinematic model of a 4 degree-of-freedom parallel mechanism for a laser weeding robot that provides high stability to the laser platform \cite{ Xuelei2016}. Xiong et al. developed a laser weeding system using a quad bike and gimbal mounted laser pointers along with a fast path-planning algorithm for high efficiency static weeding  \cite{XIONG2017494}. Rakhmatulin and Andreasen designed a compact robotic platform with a laser guiding system consisting of two motors and mirrors to hit weeds detected by a stereo camera module  \cite{agronomy10101616}. Wang et al. developed a two degree-of-freedom gimbal consisting of a 5-revolute rotational parallel manipulator for dynamic intra-row laser weeding and tested in field for higher positional accuracy \cite{9775805}. Zhu et al. designed a tracked mobile robot equipped with a five-degree-of-freedom robotic arm to precisely eliminate weeds, detected through monocular ranging, in corn fields \cite{YOLOXZhu}. Fatima et al. developed a deep learning model using YOLOv5 for identification of weeds in local crops in Pakistan which was deployed for the operation of a laser weeding robot \cite{app13063997}. 

However, laser weeding robots are still limited in terms of weeding capacity and speed, especially in rough terrains and complex calibration of weed detection algorithm and elimination mechanisms \cite{LWFOF}. In this paper, we present a six-wheeled agricultural robot with a novel double four-bar linkage suspension designed for improved stability in agriculture fields. Furthermore, instead of gimbals and robotic arms, a three-axis linear actuation mechanism is designed to guide the laser with higher precision. The robot is programmed to follow crop rows, detected by Hough Line Transform, through a feedback mechanism. Canny edge detection is used to detect crop and weed in the fields and weeds are filtered based on geometric properties such as contour perimeter and area. The performance of the robot is tested for stability, weed detection and precision of actuation mechanism for weed removal by laser exposure. Experimental results show the validity of the proposed system and effectiveness in removing both intra-row and inter-row weeds with precision. The main contributions of the work are as follows,

1. A six wheeled mobile robot is designed and tested for stability in agriculture field. It features a double four-bar linkage suspension between the front and middle wheel on both sides and a back frame connecting the rear wheels.

2. A linear actuation mechanism is designed which moves the laser with the help of lead screws in two axes and timing belt and pulley in the third axis such that laser is directly above the weed at the time of exposure.

3. The robot is tested in a set-up agriculture field to determine optimal speed, weed detection rate and operating time as well to calculate the positional accuracy and hit rate of laser actuation mechanism. 

The rest of the paper is organized as follows. Section II presents the mechanical design of the robot and laser actuation mechanism. Section III details the integration of sensors and computer vision algorithms for navigation and weed detection. Section IV elaborates the results obtained through in a set-up agriculture field pertaining to the stability of the platform along with weed detection and removal. The significance of these findings is described in section V and the paper is concluded in section VI. 

\section{Mobile Platform}
\begin{figure*}[t!]
  \centering
  \begin{subfigure}[b]{0.49\textwidth}
    \includegraphics[width=\linewidth]{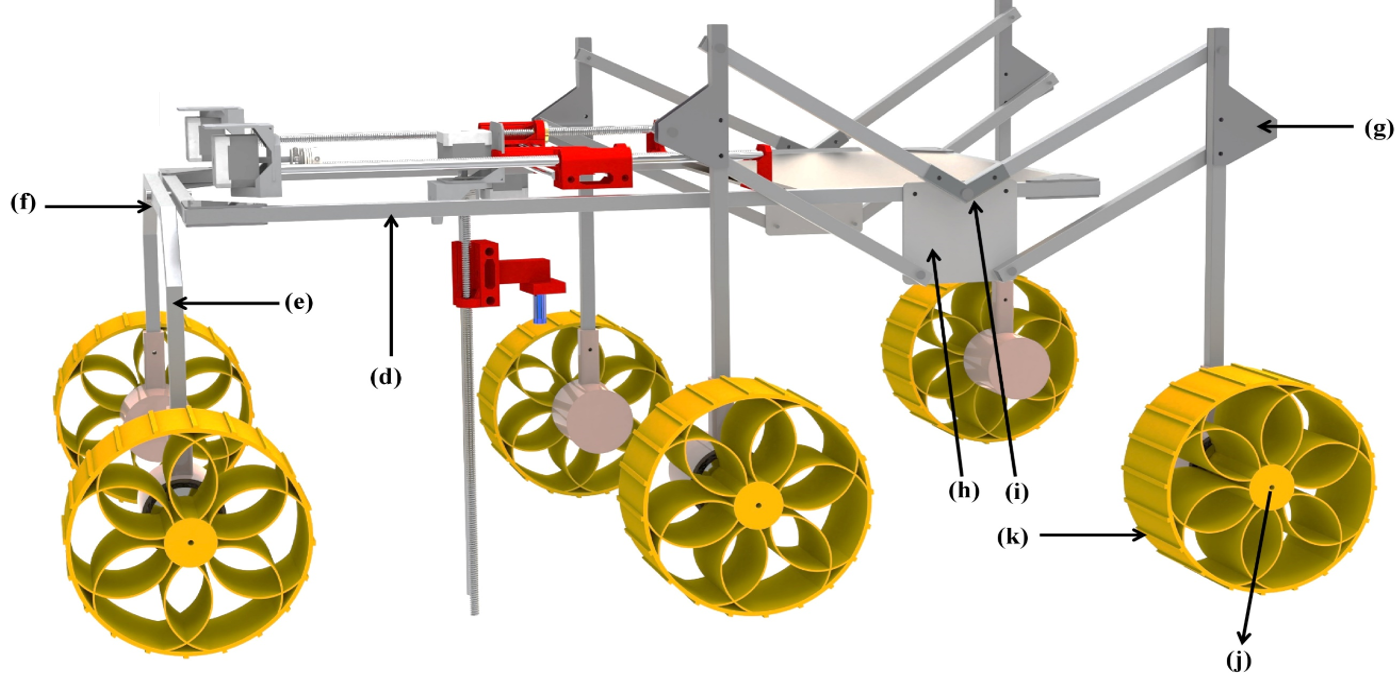}
    \caption{Side view}
    \label{fig:sideview}
  \end{subfigure}
  \hfill
  \begin{subfigure}[b]{0.49\textwidth}
    \includegraphics[width=\linewidth]{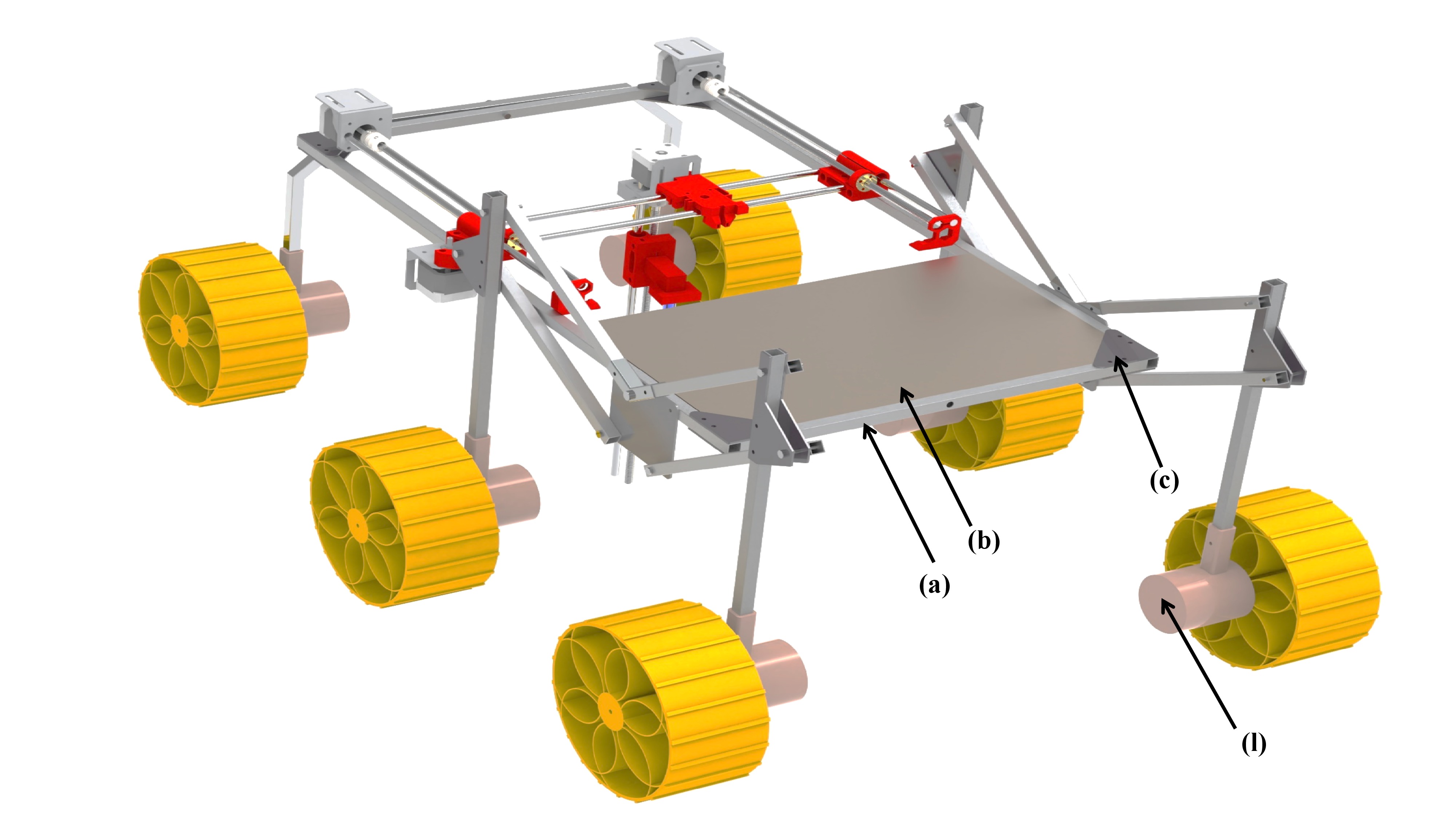}
    \caption{Front view}
    \label{fig:frontview}
  \end{subfigure}
   \caption{Design Components of Robot: (a) Chassis Front Frame, (b) Platform for Electronic Equipment, (c) Joining Bracket, (d) Chassis Side Frame, (e) Back Frame, (f) Revolute Joint between Chassis and Back Frame, (g) Bracket Connecting Vertical Link to Diagonal Links, (h) Bracket Connecting two Mechanisms, (i) Revolute Joint between chassis and common bracket (j) Custom Designed 3D Printed Wheel, (k) Rectangular Extrudes, (l) Motor Holder coupled with wheel}
  \label{fig:6wheelrobot}
\end{figure*}

A stable chassis and an effective suspension system are necessary pre-requisites for any agricultural robot to operate in a field. Furthermore, a lightweight platform is essential to minimize power consumption and soil disturbance. In order to achieve both of these objectives, a six-wheeled platform is designed and manufactured for improved stability and weight distribution. 
\subsection{Chassis}
The chassis is of rectangular shape made up of stainless steel 201 grade extrudes, joined through bolts and mild steel re-enforcements at the corners. The extrudes have a square cross section of 12.7 mm and thickness of 1.2 mm. Stainless Steel 201 grade is used in light of its high yield and shear strength (365 MPa and 292 MPa respectively) as well as machinability. The length of the frame is 760 mm and the width is 406 mm. Each extrude shares two bolts and a 2.5 mm thick mild steel plate at each end. This ensures prevention of failure due to stress concentrations under high impact scenarios and weight attached to the chassis. 

\subsection{Suspension}
A four-bar  mechanism is a modification of the four bar linkage that converts rotary motion into linear motion. A combination of four-bar  mechanisms, attached to each side of the chassis, acts as the suspension. The mechanisms share a common fixed link. Each mechanism consists of a rotating link (crank), connecting link, and a slider as shown in Fig. 2. An additional supporting link is added parallel to crank to provide structural rigidity for overcoming obstacles. 

A mild steel bracket is used as the fixed link. It is attached to the chassis frame along with the two cranks through a pin joint. The pin joint restricts five degrees of freedom but allows rotation of the cranks about the axis of the pin. The crank  of two mechanisms are welded together at an angle of $ 120^{\circ} $ to limit the maximum possible angle each crank rotates. The wheels act as the slider link. They are attached to the connecting link through a motor holder and coupled with motor shaft. The supporting link is pinned to the fixed link, joining it to the connecting link. 

The pins are locked in the lateral motion using 4 mm diameter snap locks at one end and the pin head at the other end. This ensures free rotation of suspension joints without inducing any lateral motion. When the robot encounters an obstacle, a moment force is applied that rotates the two cranks about the pin axis. As a result of this moment, the two connecting link always translate in the opposite direction. This relative motion of the two links serves as a damper to dissipate the energy of the shock. 

Similar mechanism is used in the back frame. The pivot of the back frame is pinned to the chassis frame. The pin joint allows rotation about its axis. The allowance helps to absorb the effect of obstacles and unevenness while keeping all the wheels in contact with the ground. This is important for preventing deformation of the chassis frame and ensuring stability of laser platform and image acquistion system. 
\subsubsection{Working Mechanism of Suspension}
The kinetic energy induced by a bounce is damped by utilizing the motion at the opposite ends. It provides efficient damping, necessary to keep the central platform straight. This spring-less suspension, unlike conventional systems, is suited for the platform's lightweight design. It is more effective than heavier mechanisms like rocker-bogie suspension, which are intended for larger obstacles. The double four-bar linkage suspension not only allows the platform to be light weight but also enables it to overcome sudden impacts, small slopes and obstacles.

\begin{figure*}[ht!]
  \centering
    \includegraphics[width=0.8\textwidth]{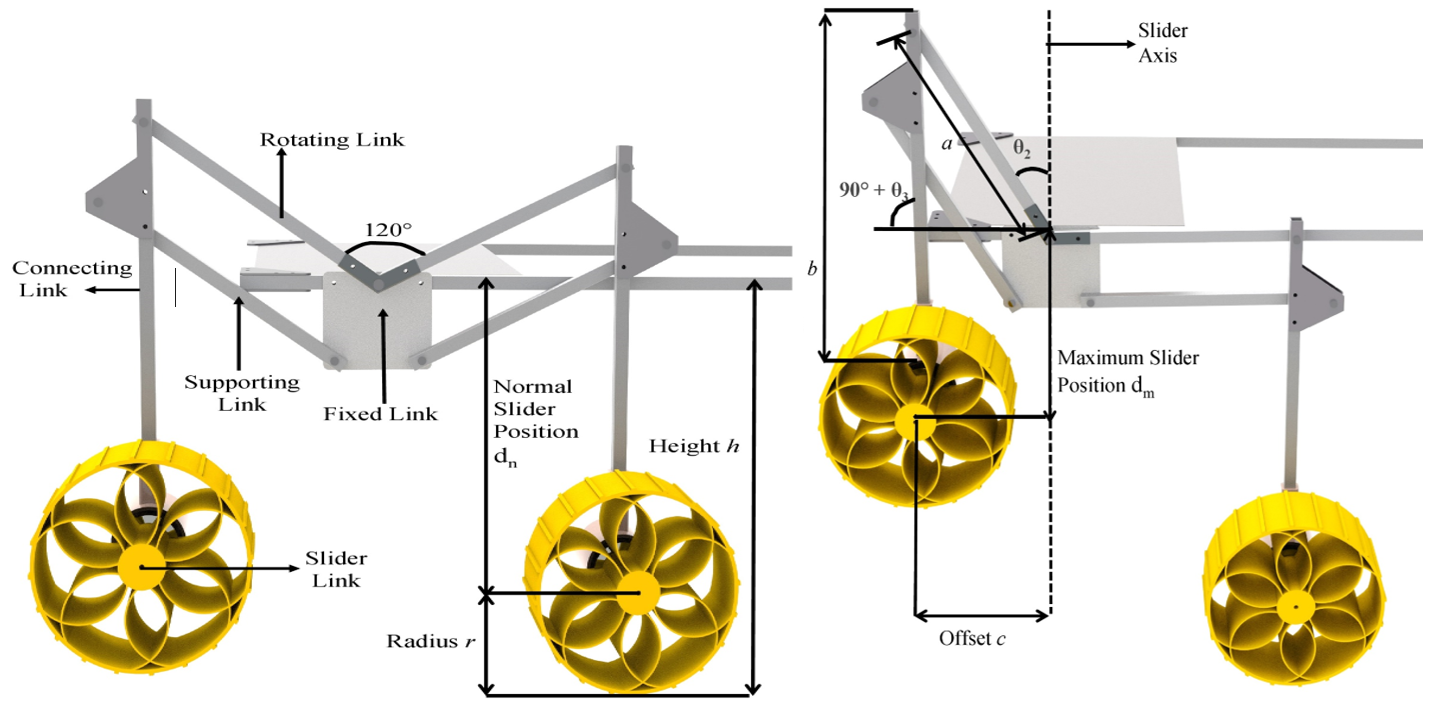} 
    \caption{Comparison of Suspension in Normal and Maximum Deflection State}
    \label{fig:suspension_comparison}
\end{figure*}

Rotation of the crank about the pin joint lifts the connecting link which causes the respective wheel to move upwards from the ground. This allows the wheel to move along the vertical axis and enables the robot to overcome obstacles. The rotation of the crank determines the distance moved by the slider which subsequently determines the size of the obstacles that can be climbed. 

The wheel of one mechanism reaches its maximum deflection when the crank of the second mechanism is aligned parallel to the horizontal chassis frame. In such a position, the rotation of the second crank is limited by the $ 120^{\circ} $ weld. This causes the wheel of the first mechanism to achieve its highest vertical position, as shown in Fig.2. By conducting a positional analysis of the four-bar four-bar  mechanism, we can determine the maximum vertical position of the slider. It is significant in predicting the largest obstacle size that the robot is capable of climbing.

\subsubsection{Determining maximum climbing ability of the robot }

\begin{table}[h]
\centering
\caption{Specifications of the Robot's Locomotion System}
\label{tab:locomotion-specs}
\begin{tabularx}{\columnwidth}{@{}Xcc@{}} 
\toprule
\makecell{Quantity} & \makecell{Denoted by} & \makecell{Value} \\
\midrule
\makecell{Crank length} & \( a \) & 270 mm \\
\makecell{Connecting link length} & \( b \) & 320 mm \\
\makecell{Fixed and slider link offset} & \( c \) & 195 mm \\
\makecell{Angle between slider axis\\ and crank} & \( \theta_2 \) & 30\(^{\circ}\) \\
\makecell{Robot Height} & \( h \) & 320 mm \\
\makecell{Wheel Radius} & \( r \) & 92 mm \\
\makecell{Original slider position} & \( d_n \) & 228 mm \\
\makecell{Maximum deflection\\ of the slider} & \( d_m \) & -90.69 mm \\
\makecell{Angle between slider axis\\ and connecting link} & \( \theta_3 \) & -10.47\(^{\circ}\) \\
\bottomrule
\end{tabularx}
\end{table}

The position of the slider link is determined with respect to the wheel's center, from where it is attached to the connecting link. The original slider position $ d_{n} $ is difference of the robot's height h and radius r as shown in fig. (2). The slider axis is vertical along which the wheel moves up and down. The difference of the original slider position and maximum slider position  gives the maximum possible lift of the wheels. 

Using vector loop method for the positional analysis of a four-bar four-bar  mechanism, the angle $ \theta_{3} $ and slider position $ d_{m} $ can be determined the following equations. \begin{equation} \theta_{3} = \sin^{-1}\left(\frac{a\sin{\theta_{2}} - c}{b}\right)  \end{equation}  \begin{equation} d_m = a\cos(\theta_2) - b\cos(\theta_3)  \end{equation} Plugging in the values in equation $ (1) $ and $ (2) $ from table .1, The angle and deflection are determined to be ,\begin{equation} \theta_3 = -10.47^{\circ} \nonumber \end{equation}\begin{equation} d_m = -90.69\text{ mm} \nonumber \end{equation}

	The angle between connecting link and vertical axis is $ 10.47^{\circ} $ indicating that the connecting link is not parallel to the slider axis in maximum deflection. The position of the slider is $ -90.67 \text{mm} $ indicating that the wheel centre is 90.67 mm below the fixed link. The maximum  theoretical deflection can be determined by the difference of original and maximum position of the slider which is equal to $ -90.67 \text{mm} $. The theoretical framework shows the maximum possible lift of the wheel by the suspension is approximately 137 mm. Thus, the maximum theoretical size of obstacle that can be climbed by the robot is 13.7 cm (5.4 inches). 
\subsection{Wheels}
The robot's wheels are designed to be flexible and self-damping through their geometry and material. The parabolic curves connecting the circumference of wheel to the shaft axis transfer the stress radially outward. This design minimizes stress concentrations near the wheel’s attachment point thus preventing shaft breakage. Additionally, rectangular extrudes on the wheel circumference enhance the contact area with the ground for optimizing traction.

The wheels are manufactured using Fused Deposition Modeling (FDM) 3D printing technology with thermoplastic polyurethane (TPU) as the material \cite{polym12112492}. Additive manufacturing provides accurate control on design of complex geometries used in the wheel. TPU ensures that each wheel has sufficient amount of elasticity, which provides a smooth drive on uneven surfaces. Furthermore, provides a damping factor of 0.25 (at ambient temperature) as well  shock absorption \cite{WU20101258}, preventing the impact from reaching the chassis and camera.

\subsection{Laser Actuation Mechanism}
\begin{figure*}[ht!]
  \centering
    \includegraphics[width=\textwidth]{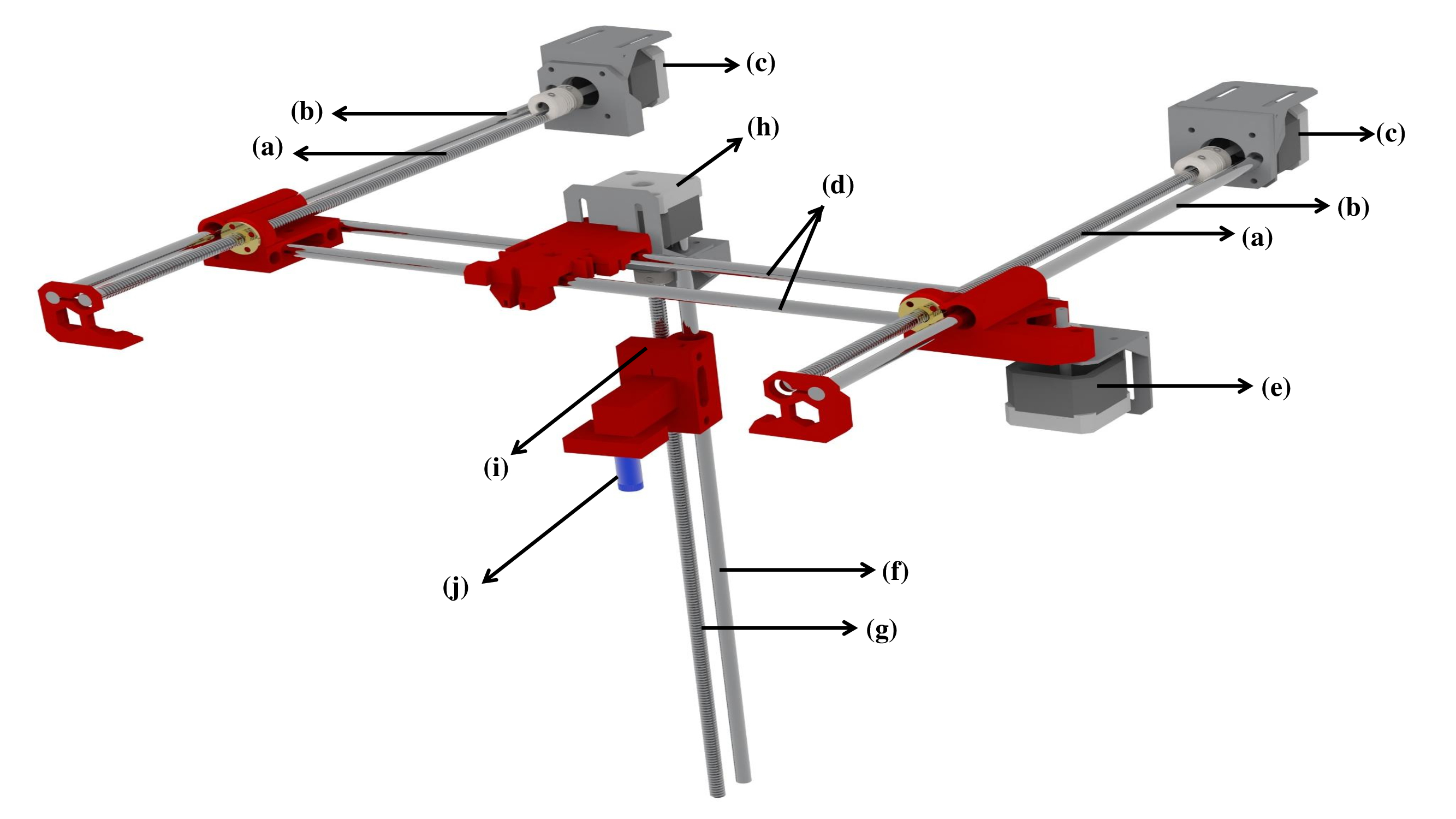}  
    \caption{Components of Laser Actuation Mechanism: (a) Y-axis lead screw, (b) Y-axis support rod, (c) Y-axis stepper motor, (d) X-axis support rod, (e) X-axis stepper motor, (f) Z-axis support rod, (g) Z-axis lead screw, (h) Z-axis stepper motor, 
(i) Laser Holding Assembly (j) Laser}
    \label{fig:laser_actuation_mechanism}
\end{figure*}
Current laser weeding robots employ either robotic arms or gimbals to direct the laser at the weeds \cite{LWFOF}. However, in our system, a simple three-axis linear mechanism is used to guide the laser to the weed coordinates. The mechanism enables vertical motion of the laser (along the z-axis) to accommodate varying weed heights. Furthermore, its linear motion across all three axes ensures that the laser is directly above the weed when it is operated. This increases the possibility of striking the meristem and burning the weed completely.  A diode laser of output power 2.5 W and wavelength of 450 nm, in the blue color spectrum, is used.  

The laser movement system is similar to the linear motion mechanisms used in CNC machines and 3D printers. Motion along x-axis is achieved by a timing belt and pulley mechanism. The timing belt is looped between an idler pulley and a tooth pulley attached to a NEMA-17 stepper motor. Controlled rotation of the stepper motor moves the 3D printed laser housing along the x-axis to the weed coordinates. The timing belt has a pitch $ p $ of 2 mm and the number of teeth $ t $ on the pulley is 20. The stepper motor has 200 steps per revolution and a micro-stepping factor $ m $ of 16 is used with the motor driver. Micro-stepping subdivides each step of the motor into finer steps, resulting in a high-resolution movement. Using the formula for resolution of motion in one step of the motor,\begin{equation} \text{Resolution }= \frac{p \times t}{N \times m} \nonumber \end{equation} The minimum distance the laser can move along x-axis is 0.0125 mm. 

Two lead screws (8 mm diameter), each coupled with a NEMA-17 stepper motor are used for motion in the y-axis. The lead screws are supported by smooth rods (also 8 mm diameter) attached to the motor holder. Synchronised rotation of the two y-axis stepper motors ensure the smooth motion of the laser assembly along y-axis along the length of the rods. A similar mechanism but with a  single lead screw is used for motion along z-axis. The laser is attached to a 3D printed housing which passes through the lead screw coupled with the motor and a smooth rod for support as shown in Fig.1. 

Since the lead screws are single thread with a 2 mm pitch, the distance (known as lead) the assembly moves in one revolution of the screw is also 2 mm. A micro-stepping factor of 16 with the inherent 200 steps per revolution yields 3200 steps within one revolution of the motor shaft. As the stepper motor and lead screw are coupled, they rotate at the same rate. This means that 3200 steps of the motor leads to a motion of 2 mm in both y and z axes. Thus, the minimum distance (in one step) which can be moved in the y and z axes is 0.000625 mm. The linear mechanism has a high resolution of 0.000625 mm in both the y and z axes, and 0.0125 mm in the x-axes making it suitable for high precision weed elimination. 

\section{System Description}
\subsection{Control Unit}
\begin{figure*}[ht!]
  \centering
    \includegraphics[width=\textwidth]{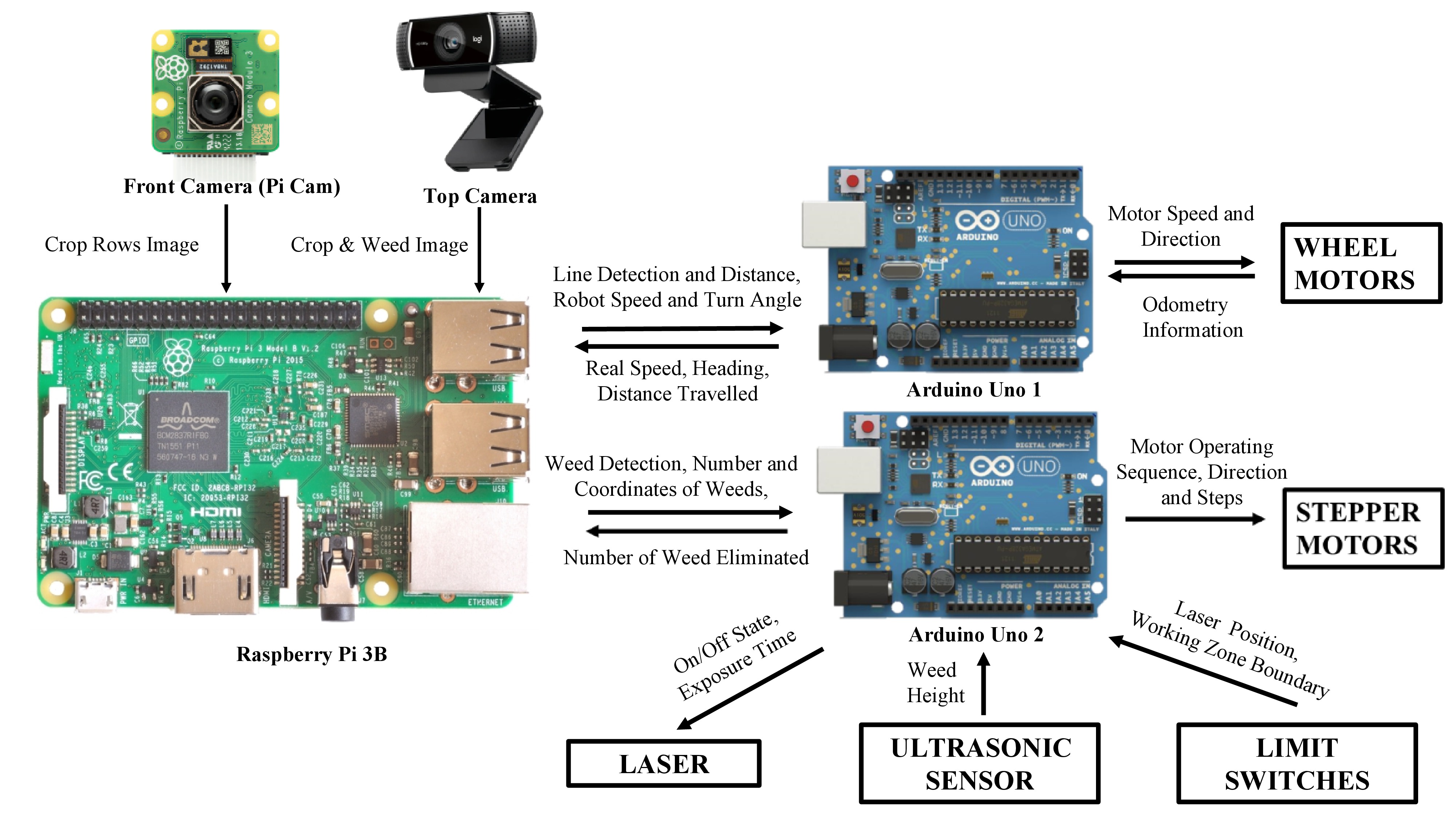} 
    \caption{Control Unit Architecture}
    \label{fig:SystemArchitecture} 
\end{figure*}
A raspberry pi 3B+ with 1 GB RAM and a 1.4 GHz processor acts as the robot's main controller along with two arduino Uno boards. One arduino controls the four stepper motors of the three-axis laser actuator and laser operation. A motor shield is attached to the second arduino for controlling the six wheel motors and receiving odometry data from the motor encoders. For visual input, the robot uses two cameras. A forward facing 5 MP raspberry pi camera captures images of crop rows. A downward facing Logitech C922x 1080p webcam, attached to the robot's rear, is used for weed detection.

The first arduino receives data from the raspberry pi about the robot's speed and navigation parameters such as distance, turn angle, and direction for aligning with crop rows. The arduino in turn controls the motors' state of rotation, speed, and direction for both straight paths and turns. If the robot deviates from the determined path, the raspberry pi signals the arduino to correct its path by changing the motor's speed and direction. The arduino also processes feedback from motor encoders to assess the robot's actual speed, heading, and distance travelled along a row and relays this information back to the raspberry Pi.

Both cameras are mounted at specific, calibrated poses. The front camera is aligned with the robot's frame and the weed detection camera is set relative to the laser's zero position. The absorption of shocks by the wheels and suspension on uneven surfaces helps maintain the pre-set pose of the cameras. Chessboard camera calibration was used for the weed detection camera such that rows and columns of the board, placed on the ground, were parallel to the chassis's frame. 

When the image processing algorithm detects a weed, the raspberry pi shares the number and coordinates of the weeds with the second arduino. The arduino in turn dictates the sequence of operation of the actuator motors as well as the direction and number of steps for each motor. The arduino then confirms with the raspberry pi to make sure all weeds have been removed. This arduino also controls the laser operation and its timing. It uses an ultrasonic sensor to measure the depth of the weed along with limit switches to keep the laser assembly within its working area.

\subsection{Navigation}
Path planning of robots in agriculture has incorporated techniques such as GPS, visual odometry, environment mapping as well as visual and laser-based Simultaneous Localization and Mapping (SLAM) \cite{8456505}.  However, visual methods of navigation for limited applications such as weeding require the robot to be guided accurately across crop rows \cite{9197114}.\cite{6907079}  developed a texture tracking method to differentiate the crop rows from soil using an overhead view of the field. \cite{VisionAsif}  used hough transform algorithm for the detection of inter-row space between the crops and calculation of the crop row pose and orientation. Additionally, \cite{8403266} demonstrated the use of hough line transform algorithm for detecting crop row patterns and measuring row angles, lateral offset as well as inter-row spacing for guidance of agriculture robots.   

For autonomous navigation in agricultural fields, the laser weeding robot uses the hough line transform algorithm. The image obtained from the front camera is converted from RGB (Red,Green,Blue) to HSV (Hue, Saturation, Value) in which crops appear as white pixels against a dark background. Noise is reduced through dilation and filtering of crop contours based on their area. Canny edge detection identifies crop boundaries, with thresholds established through multiple test iterations. 

The hough line transform algorithm returns the starting points and angles of the detected crop rows relative to the robot's current heading. In case multiple lines are detected, the one closest to the robot's centre is chosen as the crop row to be followed. The algorithm also determines the inter-row spacing required to enter the successive row. The robot then determines whether it needs to turn right or left based on the orientation of the selected line. The robot moves the necessary distance and turns by an angle corresponding to the row's angle. This alignment ensures the robot's center aligns with the center of the crop row, positioning the wheels in the inter-row spacing. Once aligned, the robot proceeds forward to detect and target weeds. 

A feedback mechanism is used to ensure the robot closely follows the crop row and the wheels do not come in contact with the crop. 
The angle of the crop row is continuously calculated and compared to the robot's heading. If there is a deviation of more than $ 5^{\circ} $ in any direction, the robot adjusts by performing a corrective turn equal to the difference between the row angle and its heading. This feedback process enables the robot to continuously realign its center with the crop row.

When the hough line detection function returns empty, indicating the end of a row, the robot exits the row to maintain a safe distance with the crops before turning. The turning maneuver consists of two separate point turns about the robot's own axis . Firstly, the robot turns $ 90^{\circ} $ in the direction of the next row. It then moves a distance equal to the inter-row spacing and turns another $ 90^{\circ} $ in the same direction. It positions itself in front of the new row to repeat the alignment and row following process. All turns are executed using skid-steering, in which the three wheels on each side rotate at equal speeds but in opposite directions, facilitating sharp point turns.

\subsubsection{Weed Detection and Elimination}
Weed detection in crops using ground based vision requires pre-processing of the image, followed by segmentation, feature extraction and classification of weeds \cite{WANG2019226}. \cite{8629137} developed a background subtraction method and morphological operation for detection of weed in early stage wheat plants. \cite{8673182} used a  combination of image segmentation, threshold-based classification, and morphological filters for precise weed identification. 

Identification of weeds for laser weeding is done based on simple geometric features. Firstly, the field image is cropped according to the dimensions of the laser's working area. It is then converted to HSV color space based on pre-set thresholds to isolate the crops and weeds from the ground. Gaussian Blur is applied to obtain a smoother image with less noise. This technique aids in edge detection by reducing the impact of irrelevant edges and highlighting the more significant edges related to weed and crop boundaries. Edge detection identifies the boundaries of crops and weeds. The detected edges undergo dilation followed by erosion. The combination of these morphological operations helps connect fragmented edges, making them more continuous and refined.

The area and perimeter of each detected contour is calculated in order to filter out the weeds. The contours having both area and perimeter below a certain percentage of the respective maximum values are disregarded as noise. Conversely, contours whose area and perimeter exceed a certain percentage of the maximum values are termed as crops. The contours lying within in these two ranges of both perimeter and area are classified as weeds. The centroid of each weed is determined by first computing the moment of each contour and then calculating the average position of all the points in the contour.

\begin{figure}[H]
\centering
\includegraphics[width=\columnwidth]{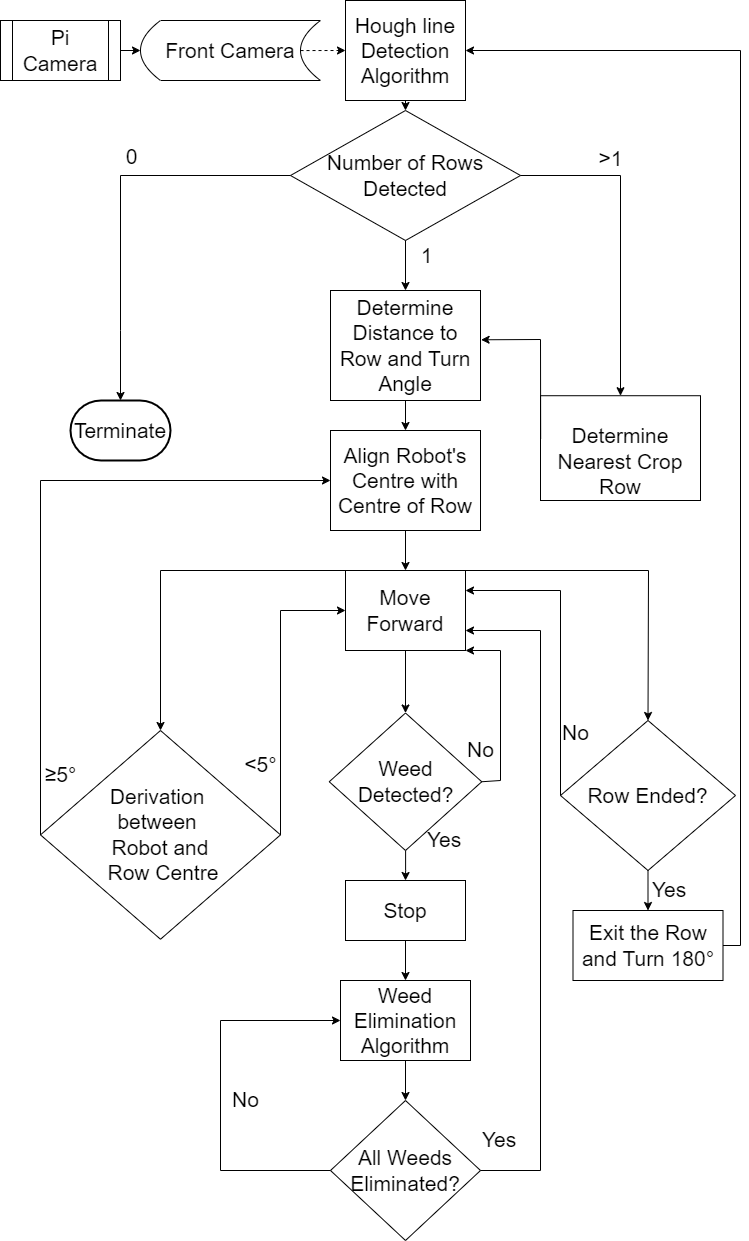} 
\caption{Methodology flowchart for crop row navigation}
\label{fig:your_label}
\end{figure}

When the robot detects a weed, it stops and sends the weeds' coordinates to the laser actuation mechanism. The laser is precisely positioned above the weed and lowered until an ultrasonic sensor detects the weed's presence. At this position, the laser is activated in proximity to the weed stem, thereby burning it completely. As the laser's output power is only 10 W, the exposure time is set to 2 seconds which is sufficient for causing permanent damage to the weeds.  

\begin{figure}[H]
\centering
\includegraphics[width=\columnwidth]{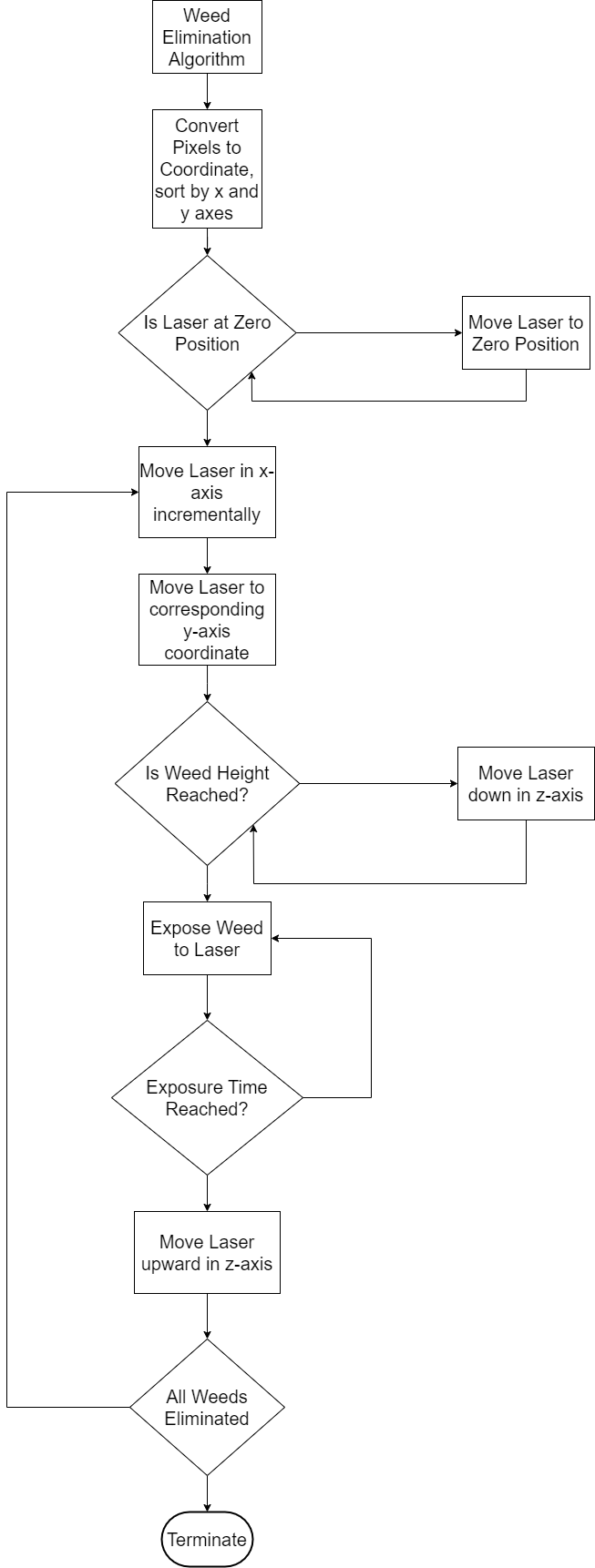} 
\caption{Methodology flowchart for weed elimination}
\label{fig:your_label}
\end{figure}

\section{Results}
The robot platform is manufactured from 201-grade stainless steel extrudes, precisely cut to fit the design dimensions. The suspension system is also integrated into the platform, attached via brackets bolted to the chassis. For the two four-bar  mechanisms, the rotating links are welded together at a  $ 120^{\circ} $ angle. Wheels are 3D printed using TPU material. Other components, such as motor holders and parts of the laser actuation mechanism, are printed with polylactic acid (PLA). Throughout the manufacturing and welding processes, tight tolerances were maintained to ensure that the final assembly closely matched the original design specifications. Except for the welded sections, all components of the robot are assembled with temporary joints, allowing for easier integration and disintegration, as well as simpler troubleshooting.
\begin{figure}[!htb] 
  \centering
  \includegraphics[width=\linewidth]{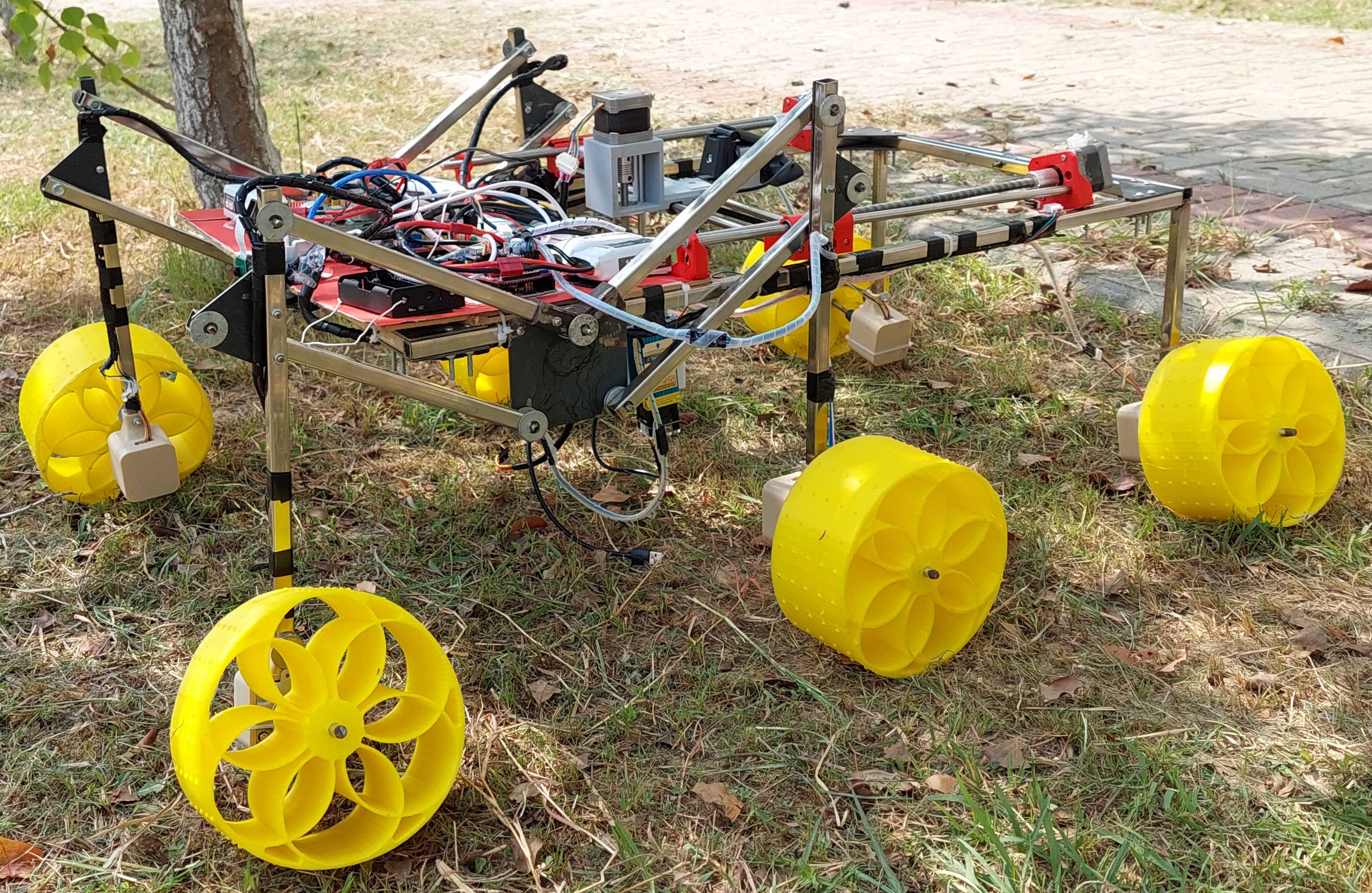} 
  \caption{Manufactured Robot Assembly}
  \label{fig:Manufactured_Robot}
\end{figure}

In evaluating the system's performance, we focused on three key factors: stability, positional error of the laser actuation mechanism and weed detection in relation to the robot's speed and weeding time. To assess stability, we placed obstacles of various sizes and shapes in the field to test the robot's ability to overcome them. These experimental results were then compared with the theoretical maximum deflection capabilities of the four-bar  mechanism. Additionally, the robot was tested at different speeds within a specified area to determine the optimal speed that balances efficient weed detection and weeding time. Lastly, we measured the positional accuracy of the laser weeding system by comparing the computed coordinates of weeds against the actual laser spot points and the extent of damage inflicted on the weeds.

\subsection{Stability}
To assess the robot's stability and its impact on navigation and image processing, a series of experiments were conducted. The robot was made to pass over various obstacles to observe its climbing ability and the effects on row-following motion and weed detection. Obstacles were classified based on size and nature: small spherical rocks ranged from 5 to 8 cm, medium-sized rocks from 10 to 12 cm, and large rocks were those bigger than 12 cm, nearing the maximum theoretical deflection of the suspension. Agricultural waste consisting of by-products and other organic matter were lumped  together in different sizes to act as potential obstacles. To simulate the effect of the robot moving up and down slopes in fields, mounds of soil were created at specific heights. The obstacle height is rounded off to the nearest unit.

In our evaluation, the robot's interaction with obstacles was categorized as 'Yes,' 'No,' and 'Partial (P).' A 'Yes' indicates successful navigation over the obstacle and continued motion, whereas 'No' denotes a failure to climb, leading to a halt. When the robot experiences some level of hindrance in overcoming the obstacle, it is referred to as 'Partial (P).' In such cases, the robot does not entirely clear the obstacle in a seamless motion but still manages to make partial progress. The impact of obstacles on the robot's navigation is also classified: 'No effect' means uninterrupted navigation; 'Light deviation' indicates a temporary path change with quick recovery through the feedback line following algorithm; and 'Significant deviation' occurs when the robot cannot resume its original path. For image acquisition, 'No effect' implies a consistent smooth image feed, 'Partial distortion' refers to brief image disturbances that quickly resolves, and 'Unstable image' describes prolonged distortion, hindering weed detection and reducing weeding efficiency.
\begin{table}[h]
\centering
\begin{threeparttable}
\caption{Effect of Different Obstacles on Robot Climbing, Navigation, and Image Feed}
\label{stability}
\small
\begin{tabular}{@{}l@{\hskip 3pt}c@{\hskip 3pt}c@{\hskip 3pt}c@{\hskip 3pt}c@{}}
\toprule
\makecell{Obstacle\\ Type} & \makecell{Size\\(cm)} & \makecell{Ability\\ to Climb} & \makecell{Effect on\\ Navigation} & \makecell{Effect on\\ Camera Feed} \\
\midrule
Small Stones & 5 & Y & NE & NE \\
Small Stones & 7 & Y & NE & NE \\
Small Stones & 8 & Y & NE & NE \\
Medium Rocks & 10 & Y & LD & NE \\
Medium Rocks & 12 & P & LD & PD \\
Large Rock & 13 & P & LD & PD \\
Large Rock & 14 & N & SD & UI \\
Large Rock & 15 & N & SD & UI \\
Agriculture Waste & 5 & Y & NE & NE \\
Agriculture Waste & 10 & Y & NE & NE \\
Agriculture Waste & 15 & P & LD & PD \\
Field Incline & 5 & Y & NE & NE \\
Field Incline & 10 & P & LD & PD \\
Field Incline & 15 & N & SD & UI \\
\bottomrule
\end{tabular}
\begin{tablenotes}
\small
\item Abbreviations: Y=yes, N=no, P=partial, NE=No Effect, LD=Low Deviation, SD=Significant Deviation, PD=Partial Distortion, UI=Unstable Image.
\end{tablenotes}
\end{threeparttable}
\end{table}

These experiments demonstrate that the robot's navigation and image quality remain stable when overcoming small obstacles. For larger obstacles within the robot's climbing ability, temporary disturbances are observed but are quickly recovered. However, obstacles exceeding the suspension's maximum deflection capacity significantly impact both navigation and image acquisition, underlining the system's limitations. Overall, the robot proves effective for use in most agricultural fields, adeptly overcoming medium-sized obstacles without compromising on navigation and weed detection.

\subsection{Determination of Optimal Speed}
The primary benefit of the weeding robot over conventional methods is the increased weed detection and reduced weeding time. Both detection efficiency and weeding duration are influenced by the robot's linear speed. Generally, increased speed should reduce weeding time, as the robot covers the same area in a shorter interval. However, higher speeds also decrease the processing time per frame, potentially leading to less accurate weed detection results. Therefore, it is necessary to operate the robot at an optimal speed that balances weed detection and weeding time. The robot's performance is the most efficient at this optimal speed. 

Given the computational constraints, the robot is better suited to operate at relatively lower speeds. In order to determine the optimal speed, the robot is operated over five different speeds ranging from 30 cm/s to 70 cm/s, with increments of 10 cm/s. Weed detection efficiency and weeding time are measured at each speed. During each trial, the robot traversed a 10-meter distance along crop rows, and the weeding time per meter was recorded, from the start of the operation to the completion of the last weed removal. To maintain consistent operating conditions, such as weed density and location, and to isolate the effect of speed on weeding efficiency, actual weed elimination was not performed. Instead, a simple laser pointer substituted the laser, simulating the weeding process without physically removing the weeds.  This approach allowed us to accurately assess the robot's weed detection efficiency and weeding time at different speeds. Weeding efficiency is measured as the percentage of detected weeds to the total number of weeds, which are manually counted before the experiment begins.

Analysis of the experimental data reveals a linear relationship with a decreasing slope between the robot's speed and both weeding time and detection efficiency. The coefficients of determination ( $ R^2 $ values) for weeding time and detection rate are 0.9821 and 0.8834, respectively. These high $ R^2 $ values indicate the reliability of the linear model for application in real field applications. The formulated linear model for weed detection percentage as a function of speed is, \begin{equation} y_1= -0.24x + 95.4 \end{equation} where  $ y_1 $ represents the weed detection percentage and $ x $ represents robot speed in cm/s. Similarly, for weeding time per meter as a function of robot speed, the linear model is \begin{equation} y_2 = -1.266x + 137.1 \end{equation} where  $ y_2 $ represents the weed time in seconds per meters.
From these models and the graph, the optimal operating speed of the robot is determined to be approximately 42.5 cm/s. This optimal speed is within a relatively slower range, which balances efficient weeding without excessively prolonging the weeding time. In subsequent experiments, the robot is operated at this optimal speed to evaluate the hit rate and positional accuracy of the laser system. 

\begin{figure*}[ht]
  \centering
  \includegraphics[width=0.8\textwidth]{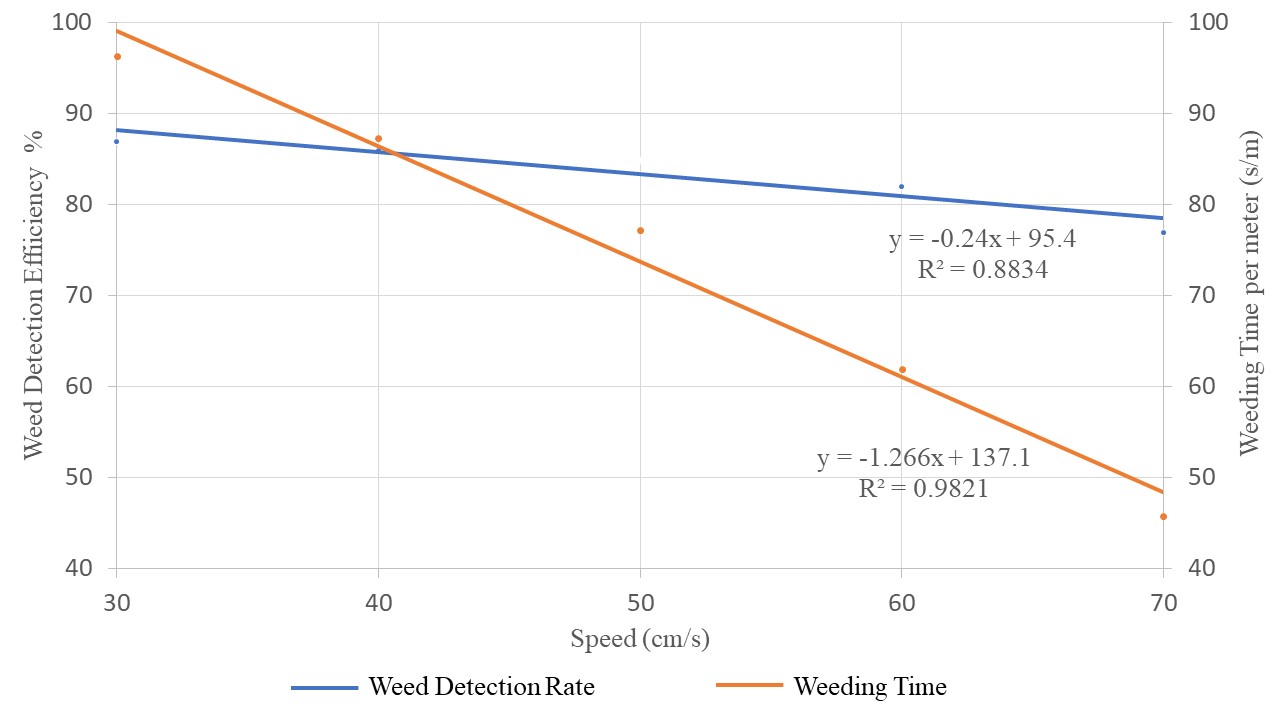} 
  \caption{Relationship between Weed Detection and Weeding Time with Robot's Speed}
  \label{fig:Graph}
\end{figure*}

\subsection{Positional Accuracy and Hit-Rate}
\begin{figure}[ht!]
    \centering
 \begin{subfigure}{0.5\textwidth}
\includegraphics[width=\linewidth]{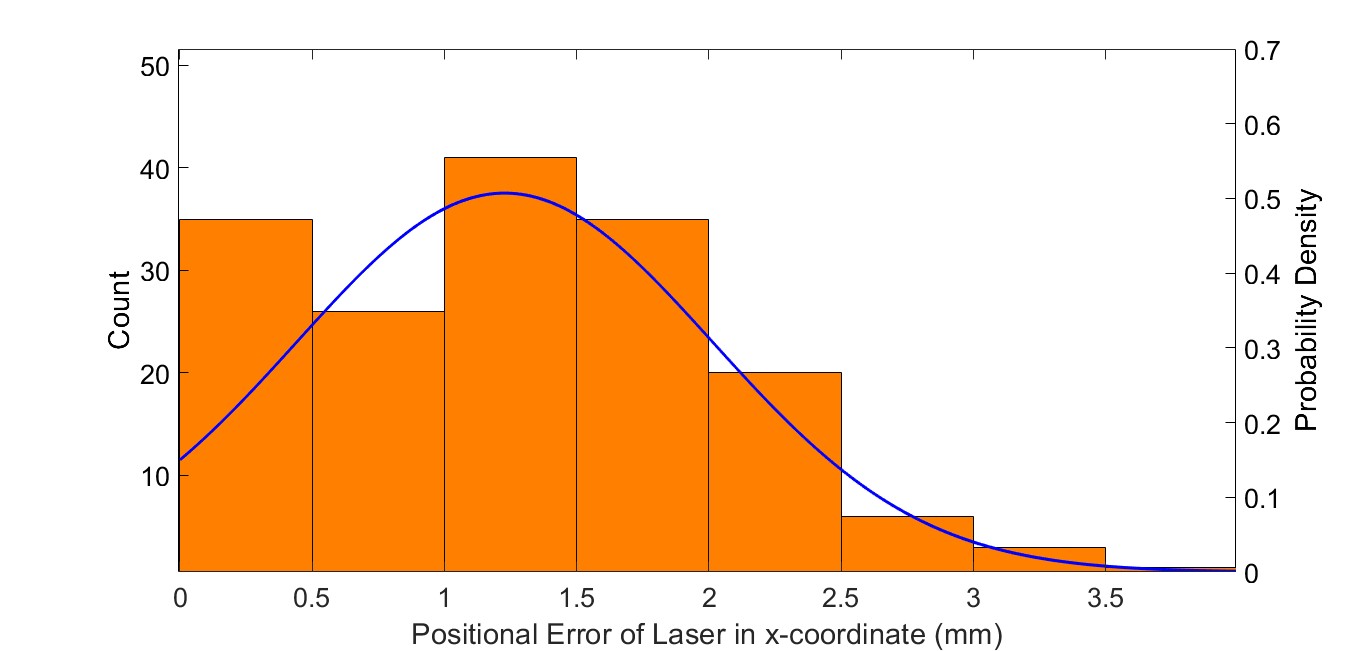}
\caption{}
\label{fig:a}
\end{subfigure}\hfill
 \vskip\baselineskip
\begin{subfigure}{0.5\textwidth}
\includegraphics[width=\linewidth]{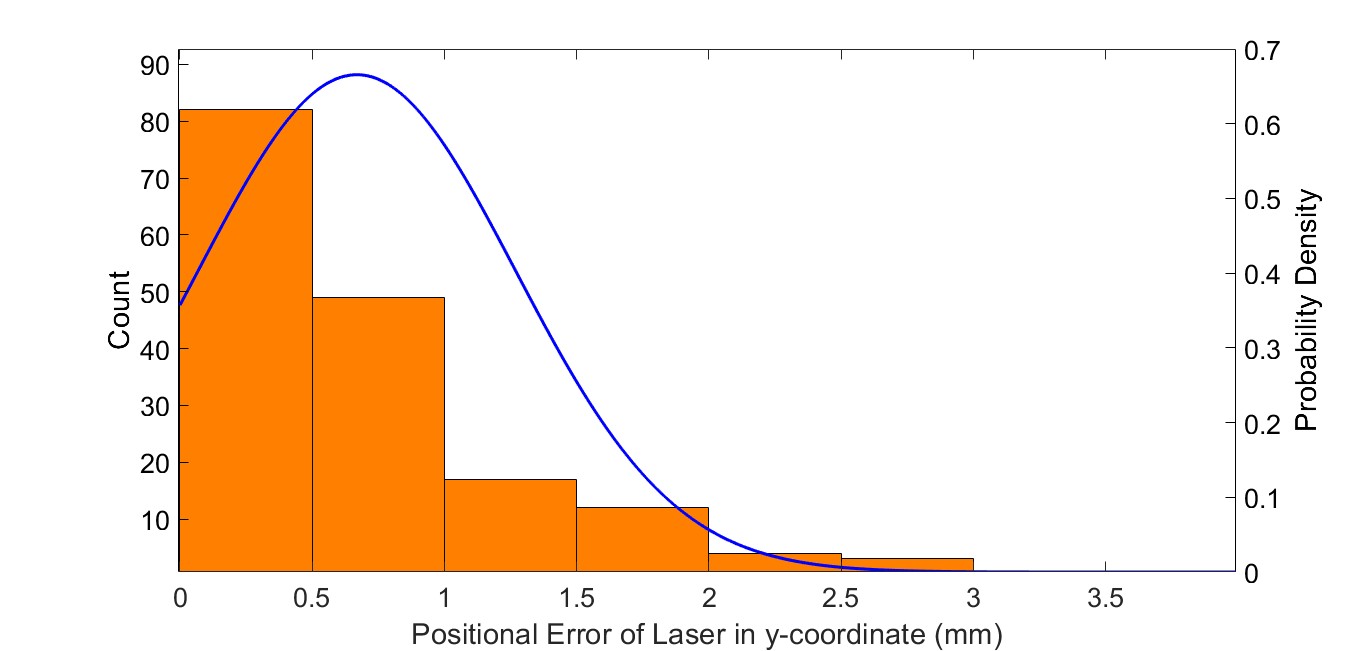}
\caption{}
\label{fig:b}
\end{subfigure}\hfill
 \vskip\baselineskip
\begin{subfigure}{0.5\textwidth}
\includegraphics[width=\linewidth]{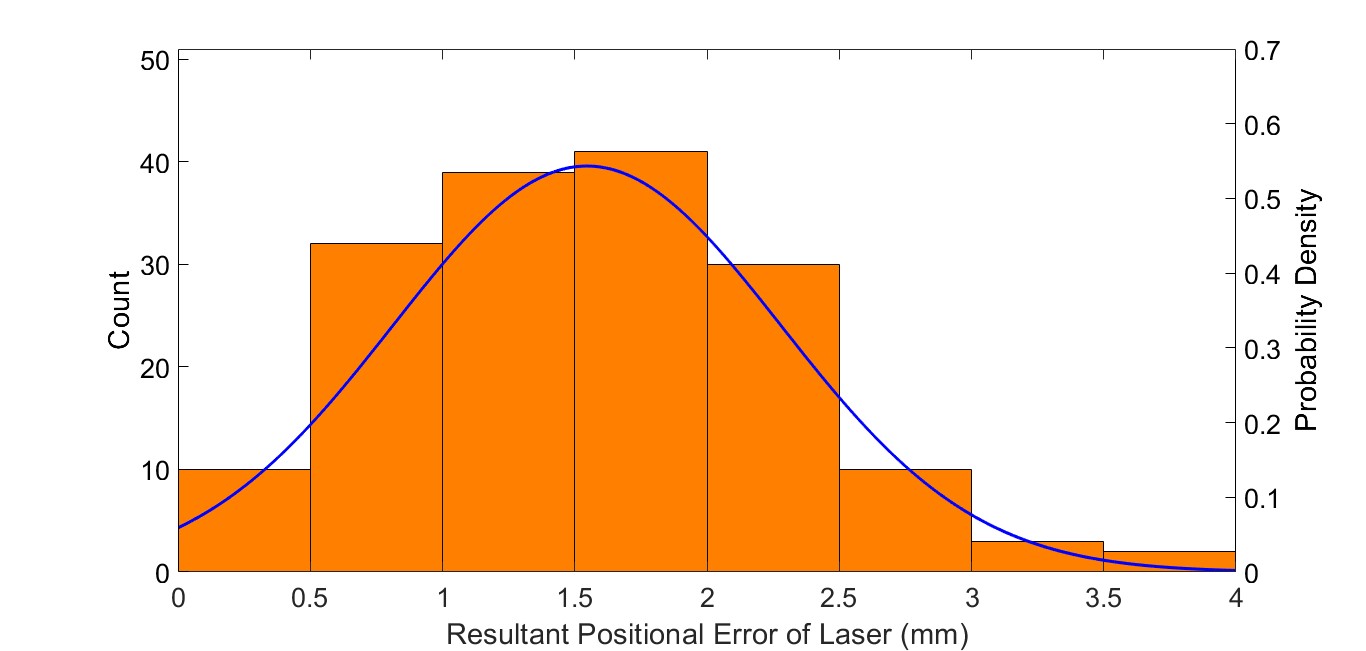}
\caption{}
\label{fig:c}
\end{subfigure}
    \caption{Histogram and Probability Distribution of Positional Errors in (a) x-axis (b) y-axis (c) perpendicular direction between weed and laser spot coordinates}
    \label{fig:errors_distribution}
\end{figure}
This paper also proposes a three-dimensional linear mechanism designed to guide the laser directly above the detected weeds. The mechanism consisting of lead screws on the y-axis and a timing belt on the x-axis, both actuated by stepper motors, moves the laser to coordinates determined by the image processing algorithm. Vertical movement of the laser (along the z-axis) is controlled based on feedback from an ultrasonic sensor which determines the weed's depth. Determining the actuation mechanism's positional accuracy on the x and y axes is essential for ensuring precise weed targeting and avoiding harming nearby crops. Moreover, the hit rate, defined as the percentage of weeds successfully eliminated out of the total detected, is also an important measure of the system’s efficacy.

In this experiment, the positional error of the laser is measured by comparing the weed's coordinates, as determined by the image processing algorithm, with the actual impact point of the laser. An additional camera, paired with the weed-detecting camera, is used to detect and calculate the coordinates of the laser spot. The two cameras are positioned directly opposite each other, producing images that are effectively vertical mirrors of one another. The image processing algorithm of the secondary camera only detects color equivalent to the 450 nm wavelength of the laser, corresponding to RGB values of (0,70,255).  The resulting image shows a circular blob representing the laser's impact point. By computing the centroid of this laser spot and applying a coordinate transform between the two images, the error in the x and y coordinates is determined for each weed. Moreover, the perpendicular distance between the weed and laser spot centroids is also determined which is the resultant of the vector difference between the coordinates of the two centroids. A weed is considered eliminated if the laser burns major parts, such as the meristem, thereby preventing its further growth and nutrient utilization.  

The robot is operated at an optimal speed of 42.5 cm/s, as determined in the earlier experiment. A manual count identified 193 weeds, with 167 detected by the robot, yielding an efficiency of 86.5\%. The mean positional errors on the x and y axes, and the resultant error, were 1.22 mm, 0.67 mm, and 1.54 mm respectively, with corresponding standard deviations of 0.79, 0.60, and 0.73. The distribution of the errors and their probability density is shown in Fig. 9. The highest frequency of positional errors along the x-axis occurs within the 1 to 1.5 mm range, whereas y-axis errors are predominantly between 0 and 1 mm. There is a significant reduction in the number of errors exceeding 2 mm along the x-axis and 1 mm along the y-axis. The probability density curves show that errors of 1 to 1.5 mm on the x-axis and 0.5 to 1 mm on the y-axis are the most likely to  occur. For resultant positional errors, a uniform distribution extends to 2.5 mm, beyond which there is a gradual decline, with the highest probability occurring at 1.5 mm.

\begin{figure*}[ht!]
    \centering
    \begin{subfigure}{0.32\textwidth}
        \includegraphics[width=\linewidth]{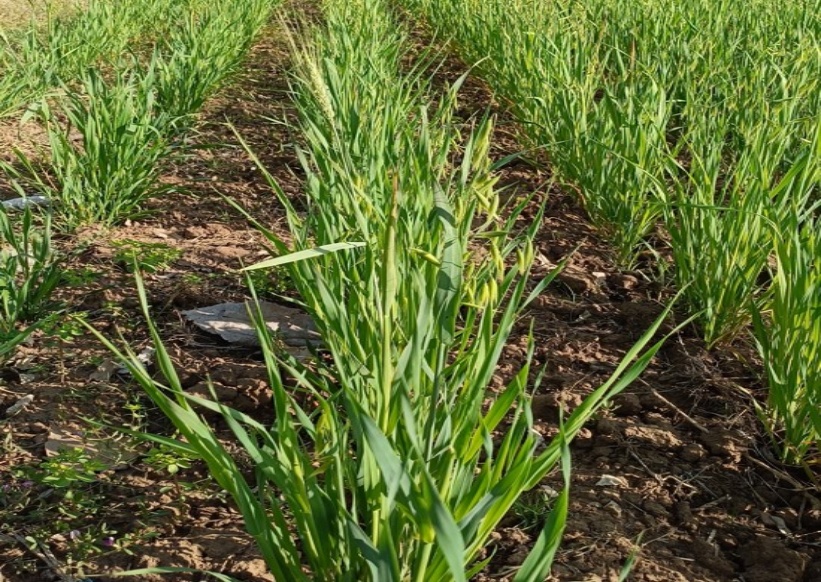}
        \caption{}
        \label{fig:1a}
    \end{subfigure}
    \hfill
    \begin{subfigure}{0.32\textwidth}
        \includegraphics[width=\linewidth]{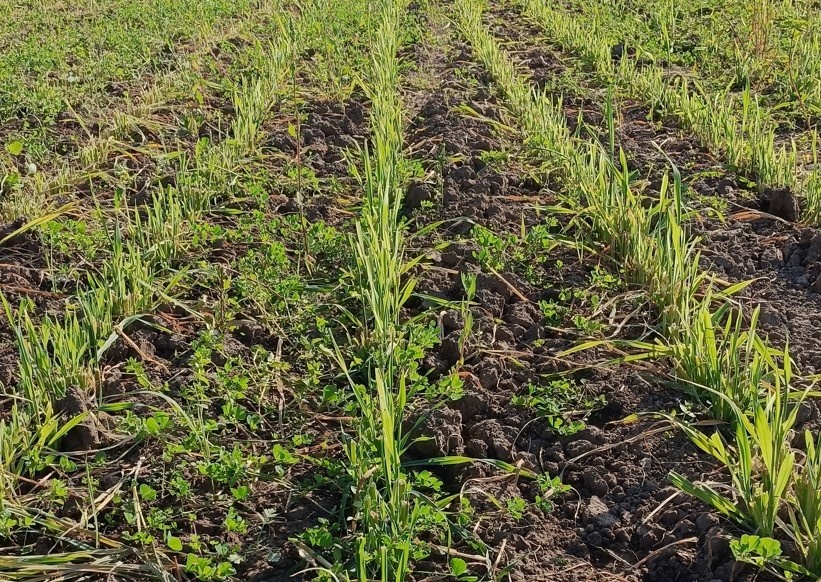}
        \caption{}
        \label{fig:1b}
    \end{subfigure}
    \hfill
    \begin{subfigure}{0.32\textwidth}
        \includegraphics[width=\linewidth]{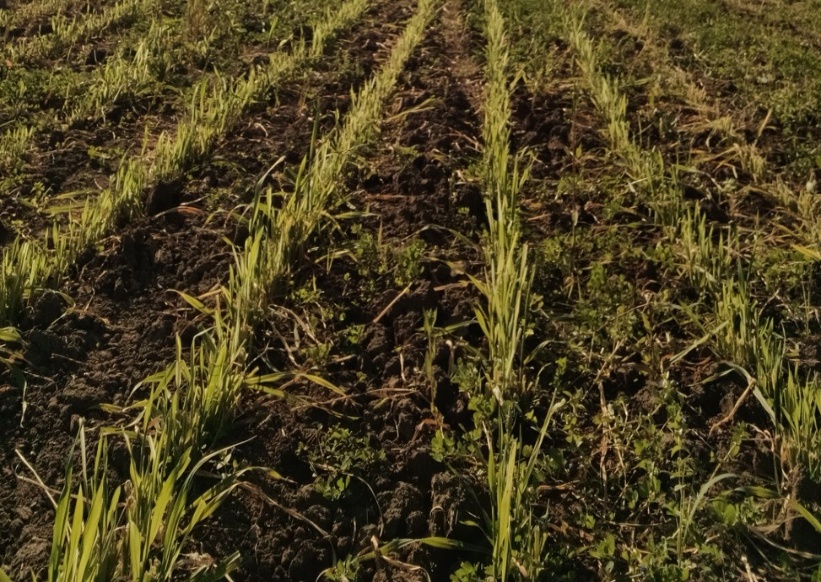}
        \caption{}
        \label{fig:1c}
    \end{subfigure}
    \vskip\baselineskip 


    \begin{subfigure}{0.32\textwidth}
        \includegraphics[width=\linewidth]{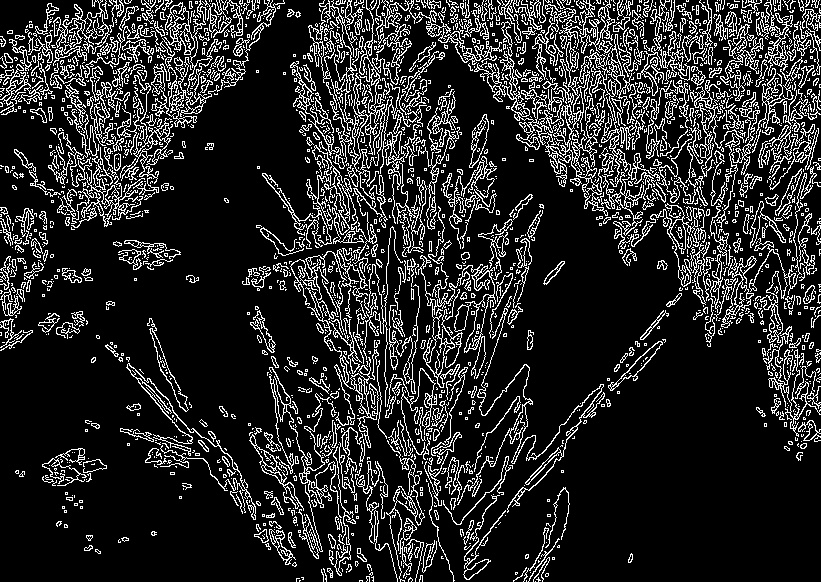}
        \caption{}
        \label{fig:3a}
    \end{subfigure}
    \hfill
    \begin{subfigure}{0.32\textwidth}
        \includegraphics[width=\linewidth]{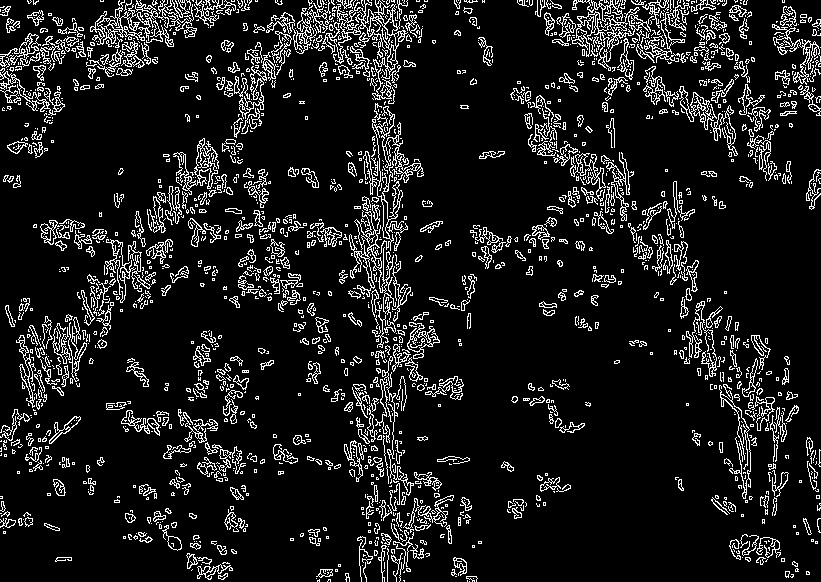}
        \caption{}
        \label{fig:3b}
    \end{subfigure}
    \hfill
    \begin{subfigure}{0.32\textwidth}
        \includegraphics[width=\linewidth]{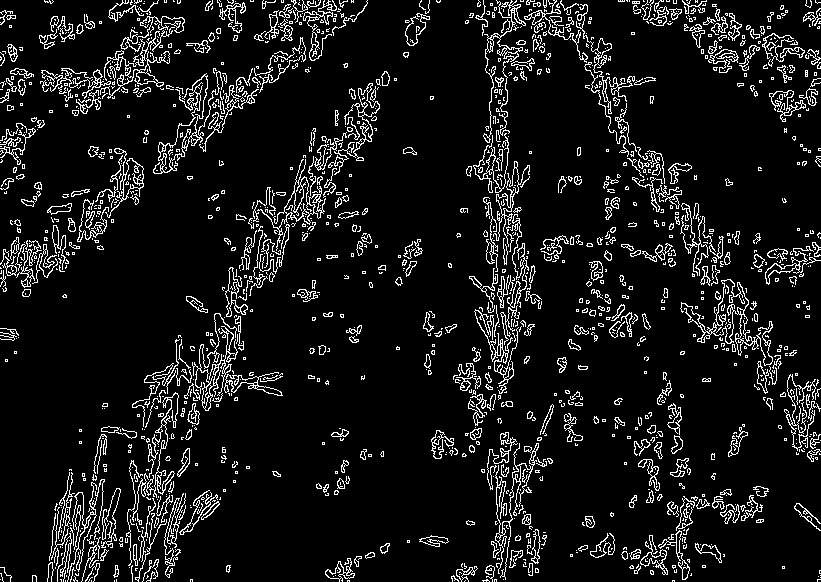}
        \caption{}
        \label{fig:3c}
    \end{subfigure}
    \vskip\baselineskip 

    \begin{subfigure}{0.32\textwidth}
        \includegraphics[width=\linewidth]{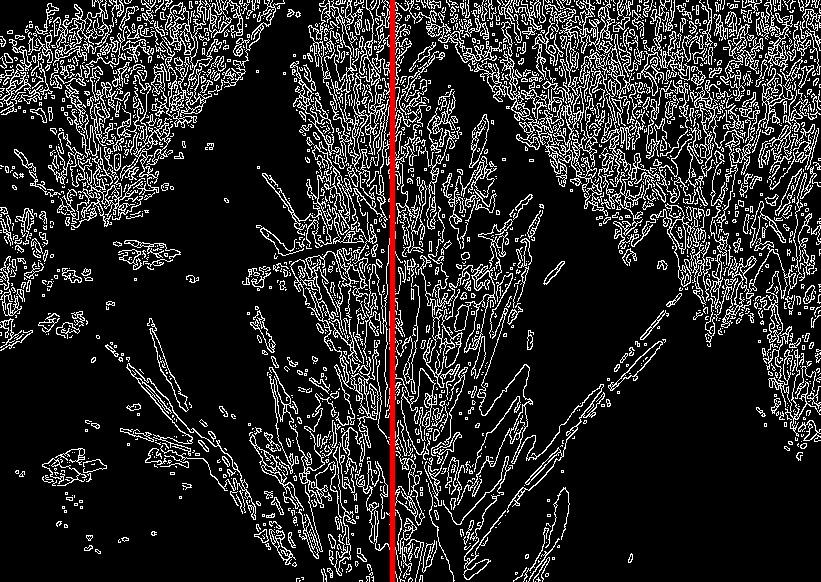}
        \caption{}
        \label{fig:4a}
    \end{subfigure}
    \hfill
    \begin{subfigure}{0.32\textwidth}
        \includegraphics[width=\linewidth]{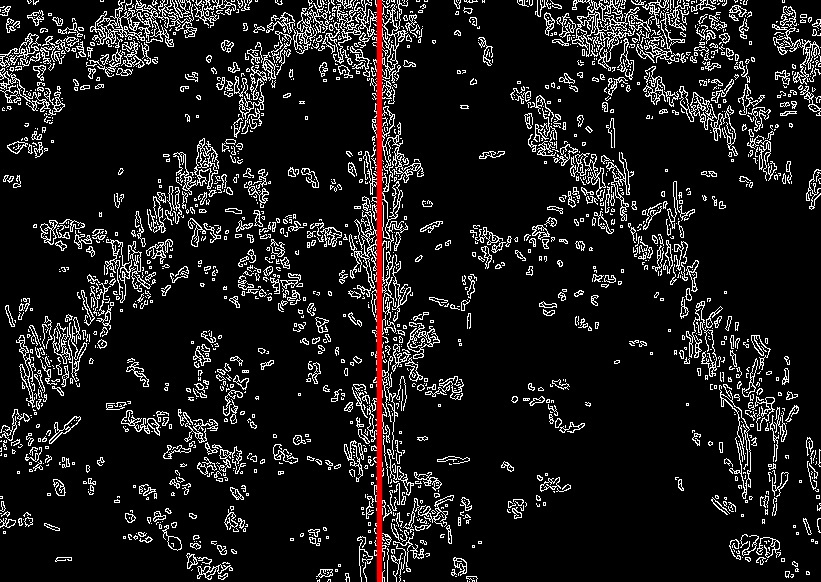}
        \caption{}
        \label{fig:4b}
    \end{subfigure}
    \hfill
    \begin{subfigure}{0.32\textwidth}
        \includegraphics[width=\linewidth]{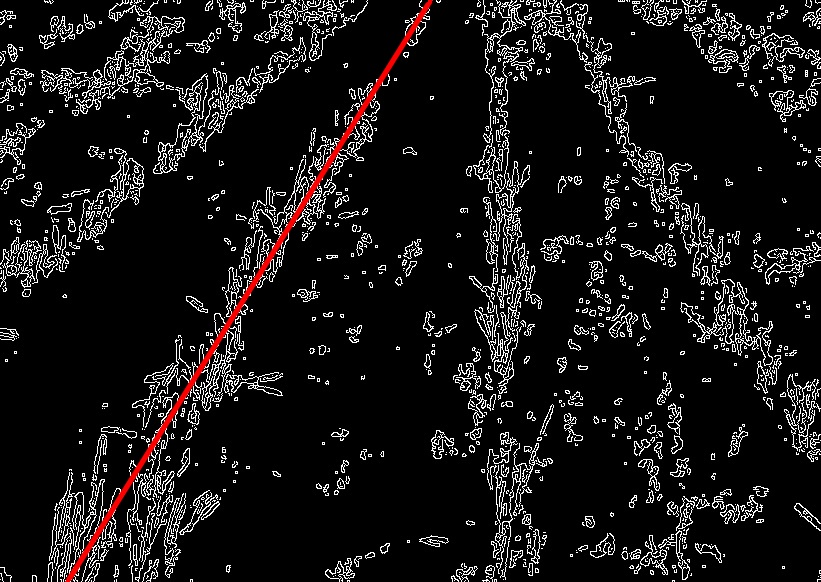}
        \caption{}
        \label{fig:4c}
    \end{subfigure}
    
    \caption{Crop Row Detection and Following through Hough Line Transform a) b) c) Original Camera Image d) e) f) Edge Detection g) h) i) Hough Line Transform and Path Detection.}
    \label{fig:grid}
\end{figure*}

\section{Discussion}
\subsection{Analysis of Suspension}
Stability is crucial for effective operation of the laser weeding robot, especially as it traverses diverse agricultural terrains. Agriculture robots encounter various obstacles, which can not only distort image acquisition and affect weed detection but also potentially impede motion. To address this, the robot is equipped with double four-bar  mechanisms on each side, coupled with a back frame. This suspension design enables the robot to smoothly navigate different terrains and overcome obstacles without the risk of stopping or toppling over. The suspension also dampens shocks and impacts, preventing them from distorting image acquisition, thereby improving weed detection efficiency. In addition, improved stability contributes to maintaining straight-line motion along crop rows and ensures precise laser positioning for accurate weed targeting.

The experimental results indicate that the robot easily climbs stones and rocks up to 10 cm in height. As the size of the obstacle nears the theoretical maximum lift of 13.7 cm, the robot still manages to overcome them but with noticeable difficulty.  The robot is unable to overcome any obstacles greater than the theoretical threshold. Agricultural waste can be traversed by the robot even when it is above 13.7 cm in height, likely due to its less uniform structure compared to rocks. Vertical inclines present a more difficult challenge. The robot is able to overcome inclines up to 10 cm with some difficulty but is unable to climb inclines higher than 15 cm. 

The obstacle climbing ability of the robot aligns with the theoretical prediction, although the suspension's performance varies depending on the obstacle type. The maximum obstacle size that the robot overcomes smoothly is slightly smaller than the theoretical limit. However, the robot is still capable of partially overcoming obstacles up to this limit. This decrease in the climbing ability and instances of partial climbing can be attributed to loss of traction between the wheels and the ground as well as discrepancies between manufacturing and assembly compared to the initial design. Despite these practical limitations, the suspension's theoretical model is validated by the experimental performance to an appreciable degree. 

Small rocks have negligible impact on the robot's navigation and image quality. As the size of the rocks increases, there is a significant increase in deviation from the desired path. Once the size exceeds the theoretical limit, the robot fails to recover its path while attempting to climb these obstacles. On the contrary, organic obstacles, even slightly larger in height, do not significantly impact the robot's path planning. However, when encountering vertical inclines, the robot's navigation is slightly disturbed at around 10 cm and becomes increasingly unstable at heights above this.

Image quality and weed detection are minimally affected by small rocks and organic matter. But, as the size of the obstacle increases and the robot's climbing ability is reduced, it results in the transfer of vibrations and impact energy across the platform. While the wheels and suspension dampen these shocks and impacts to some extent, image quality is ultimately distorted at larger obstacle heights, hindering weed detection. However, for medium-sized obstacles that the robot can fully or partially climb, the suspension efficiently protects the image quality.

The obstacle climbing ability of six-wheeled autonomous rovers, such as those with the rocker-bogie mechanism, depends on wheel diameter and overall rover dimensions. A rocker-bogie mechanism of comparable size can climb obstacles up to around 15 cm \cite{RockerBogie}. All-terrain vehicles (ATVs), despite their high maneuverability, are often heavyweight and cause significant soil disturbance. In this context, the performance of our suspension system is comparable with state-of-the-art designs. The proposed suspension has an advantage over its contemporaries of being springless, lightweight, and easy to manufacture and assemble. 

\subsection{Weed Detection and Weeding Time}
The optimal performance of the robot is a trade-off between weed detection efficiency and weeding time. Higher performance of these two parameters is essential for the robot to be an effective alternative to manual weeding. The developed linear models for detection efficiency (eq.3) and weeding time (eq.4) demonstrate a linear relationship with speed, depicting a decrease in both quantities as speed increases. The intersection of these models suggests that the robot's best performance occurs at an optimal speed of 42.5 cm/s. The data indicates that while weeding time decreases with an increase in speed, detection efficiency similarly declines. It's important to note that the reduction in weeding time is not only due to the increased speed but also results from the decreased number of weeds to be eliminated.

As the robot's speed increase from 30 cm/s to 50 cm/s, the detection rate decreases very gradually from 87\% to 85\%. However, upon increasing the speed beyond 50 cm/s, there is a noticeable decline in weed detection efficiency to 77\% at a speed of 70 cm/s. This pattern indicates that the detection rate decreases more steadily at higher speeds, while at lower speeds, there is no significant change. We can conclude that effect of robot speed on weed detection only becomes prominent at speed higher than 50 cm/s. Therefore, the weeding robot is best suited to be operated at slower speeds. 

The weeding time per meter is affected more significantly by the reduction in the number of weeds detected than by the increase in the robot's speed. As the robot's speed increases, fewer weeds are detected, which in turn, leads to a reduction in weeding time. This reduction in weeding time with increased speed follows a trend similar to the weed detection rate. Up to a speed of 50 cm/s, the decrease in weeding time is gradual yet more prominent than the decrease in detection rate. However, beyond 50 cm/s, the decline in weeding time becomes much steeper.

The optimal speed estimated from the experimental data and linear models of weeding time and detection rate is 42.5 cm/s. Operating below this speed ensures high weed detection but results in prolonged weeding operation. Conversely, exceeding the optimal speed decreases the weeding time but at the cost of weed detection accuracy. Operating the robot at the optimal speed makes it possible to achieve faster operation without compromising on the weed detection efficiency. The robot demonstrates notable performance in detecting weeds and covering larger fields in shorter times, making it suitable for sustainable and organic automated weeding in real agricultural applications. 

\subsection{Accuracy of Linear Actuation Mechanism}
Laser positioning along the y-axis exhibits a lower mean error of 0.67 mm with a standard deviation of 0.60, compared to the x-axis, which has a mean error of 1.22 mm with a standard deviation of 0.79. Additionally, a higher concentration of y-axis errors is observed in the lower range compared to the x-axis, as depicted in the bar charts and probability distributions in Fig. 9(a) and 9(b). The higher accuracy along the y-axis is attributed to the finer resolution of its lead screw mechanism, 0.000625 mm, compared to the x-axis's timing belt at 0.0125 mm. The resultant positional error combines the individual errors of the x and y axes and is thus greater than the individual errors. 

The resultant error between the weed centroid and laser spot represents the true positional accuracy of the three-axis linear mechanism. This perpendicular distance between the two centroids is the aggregate of the individual errors along the x and y axes. The resultant positional error is 1.54 mm with a standard deviation of 0.53. The errors are uniformly distributed between 0 and 4 mm, with the highest probability of error occurring at 1.5 mm and the majority of errors falling within the 1 to 2 mm range. The resultant error exhibits a bell curve trend more clearly than the other two errors, as depicted by the probability distribution curve in Fig. 9(c). Since the mean resultant error of 1.54 mm is close to the highest probability of error occurrence at 1.5 mm, we can conclude that the resultant positional error more closely approximates a normal distribution than the individual errors.
 
The laser weeding robot achieved an impressive hit rate of 97\%. Accurate positioning of the laser, particularly along y-axis helps in permanently eliminating majority of the weeds detected. In addition to accurate positioning, the mechanism's ability to adjust the laser's vertical position depending on the weed height, using feedback from an ultrasonic sensor significantly contributes to its success. This adjustment ensures that the laser's working area remains optimally close to the target, preventing the dispersion of the laser spot and ensuring that the thermal energy is precisely focused on the weed. As a result of these factors, the laser can effectively target the weeds, leading to a successful hit rate of 97\%. In the study by  \cite{XIONG2017494}, the gimbal-mounted laser system had a mean error of 1.97 mm, whereas our three-dimensional linear actuation mechanism demonstrated a lower mean error of 1.54 mm over a larger number of weeds. In addition, the motion of laser in z-axis helps achieve a similar hit rate of 97\%. Thus, the three-dimensional linear weeding mechanism can be applied to field application with low positioning errors and high weed elimination rate.

\section{Conclusion}
In this study, a laser weeding robot system equipped with a double four-bar linkage suspension and a three-dimensional laser actuator mechanism was designed and manufactured. The robot was tested in an outdoor agricultural field setting to assess the stability of the novel suspension mechanism, determine the optimal operating speed, and evaluate the positional accuracy and hit-rate of the weeding mechanism. 

The robot is designed for enhanced stability on rough agricultural terrain. The front and middle wheels are connected through a combination of four-bar mechanisms for climbing obstacles. The back frame links the rear wheels, allowing lateral movement to dissipate shocks. Positional analysis of the suspensions provides a theoretical limit of 13.7 cm as the maximum size of obstacles the robot can climb. In field tests, the robot managed to overcome organic matter up to 15 cm and rocks up to 10 cm. However, it struggled with obstacles around 13 cm, slightly affecting its navigation and image quality. The performance decreased on vertical climbs, managing inclines of 5 cm well but unable to handle 15 cm, significantly affecting motion and image processing. As long as the robot was able to overcome obstacles, the impact on motion and image analysis did not hinder its ability to detect weeds and navigate. The robot's flexible wheels, 3D-printed from TPU, absorbed shocks, contributing to stable image capture and laser positioning.

The control unit of the robot is a Raspberry Pi 3B, which controls two Arduinos responsible for the motors of the wheels and the laser mechanism, with data communication via ROS nodes. Computer vision algorithms, implemented using the OpenCV library, are utilized for detecting crop rows and weeds. Crop rows are detected and followed using the Hough Line Transformation to ensure that the robot’s center aligns with the center of the crop row. Weeds are differentiated from crops using contour detection and geometric features. Performance tests at speeds from 30 to 70 cm/s revealed that weed detection efficiency and weeding time both depend on speed, displaying linear trends with decreasing slopes. At the optimal speed of 42.5 cm/s, the robot achieves an 86.5\% weed detection rate with a weeding time of 85 seconds per meter, making it a promising alternative to manual weeding.

The laser weeding mechanism uses lead screws on the y and z axes and a timing belt on the x-axis, controlled by stepper motors to precisely target weeds. The laser's position is adjusted perpendicularly above each detected weed and is lowered towards it based on its height, determined by an ultrasonic sensor. This precise three-axis movement, involving a 10 W, 450 nm wavelength laser operated for 2 seconds, results in a 97\% weed elimination hit rate. The system achieves low positional errors: 0.67 mm in y-axis and 1.22 mm in x-axis with respective standard deviations of 0.6 and 0.79. The resultant positional error of 1.54 mm marks an improvement over the 1.97 mm mean error reported in the study by \cite{XIONG2017494}, while maintaining a similar hit rate. This three-axis linear mechanism represents an advancement in robotic laser weeding, offering an effective, chemical-free farming solution. 

However, the robot is not without its limitations. Its limited computational power restricts the use of advanced algorithms that could further enhance performance and efficiency. The crop detection system guides the robot but lacks real-time field positioning and obstacle detection. It could be improved by implementing visual SLAM but is currently too computationally expensive for the system \cite{agrislam}. The weed detection method works well for specific weeds but lacks flexibility and adaptability across different crop types and weed species, highlighting the need for deep learning techniques \cite{HASAN2021106067}. Additionally, the 10W laser, while effective, slows down weeding over larger areas; a higher power laser could significantly reduce weeding time. Future prospects include increasing the robot's overall dimensions to climb bigger obstacles and integrating deep learning for better weed recognition. Implementing SLAM for field navigation along with a weed damage model tailored to weed size and laser power could optimize weeding time and power consumption, making the robot more viable for extensive agricultural use.

\section*{Conflicts of Interest}
The authors declare that there are no conflicts of interest regarding the publication of this paper.

\section*{Acknowledgements}

\bibliographystyle{IEEEtran}
\bibliography{bibfile}

\end{document}